# Vision-Cloud Data Fusion for ADAS: A Lane Change Prediction Case Study

Yongkang Liu, *Student Member, IEEE*, Ziran Wang, *Member, IEEE*, Kyungtae Han, *Senior Member, IEEE*, Zhenyu Shou, Prashant Tiwari, *Member, IEEE*, John H.L. Hansen, *Fellow, IEEE*

*Abstract*—With the rapid development of intelligent vehicles and Advanced Driver-Assistance Systems (ADAS), a new trend is that mixed levels of human driver engagements will be involved in the transportation system. Therefore, necessary visual guidance for drivers is vitally important under this situation to prevent potential risks. To advance the development of visual guidance systems, we introduce a novel vision-cloud data fusion methodology, integrating camera image and Digital Twin information from the cloud to help intelligent vehicles make better decisions. Target vehicle bounding box is drawn and matched with the help of the object detector (running on the ego-vehicle) and position information (received from the cloud). The best matching result, a 79.2% accuracy under 0.7 intersection over union threshold, is obtained with depth images served as an additional feature source. A case study on lane change prediction is conducted to show the effectiveness of the proposed data fusion methodology. In the case study, a multi-layer perceptron algorithm is proposed with modified lane change prediction approaches. Human-in-the-loop simulation results obtained from the Unity game engine reveal that the proposed model can improve highway driving performance significantly in terms of safety, comfort, and environmental sustainability.

## I. Introduction and Background

According to a survey from the National Highway Traffic Safety Administration (NHTSA), human errors contribute to 94% of road accidents [1]. With the goal of improving road safety and efficiency, enormous efforts have been made towards the development of electrical vehicles [2], intelligent transportation systems [3], and fully autonomous vehicles [4]. However, a large number of challenges present in this process because of the complex nature of driving, as well as factors such as system capability, manufacturing costs, administration regulations, and community safety. It is expected that during the transition period to full automation, vehicles with different levels of automation and human driver engagement will be mixed together to form large-scale traffic environments. Therefore, to create a safe and efficient driving experience, visual guidance systems are suggested/necessary to alert potential risks and/or provide essential advice (e.g., lane level guidance) to drivers.

Environment perception has been an essential component for both intelligent vehicles and Advanced Driver-Assistance Systems (ADAS). In addition, recent advancements in signal processing for smart vehicle technologies suggest that there is a growing need to better integrate drivers and vehicles in the research design of future vehicles [5], including conversational vehicle-driver interaction [6]. International collaborations (Japan, USA, Turkey) suggest that cooperation between groups could accelerate best practices [7], as well as advance driver modeling and system development for distraction detection [8]. Common applications, such as object detection, semantic segmentation, drivable area detection, and object tracking, have been researched for decades in the field of computer vision with cameras as the most commonly used sensors. However, only real-time information of neighboring vehicles can be obtained based solely on cameras and state-of-the-art computer vision algorithms. Predicting behaviors of a target vehicle is difficult because of the short advance detection time and the lack of historical data.

One recent emerged representation of Internet of Things (IoT) is the Digital Twin technology, which has the capability of assisting prediction by providing historical data stored in the cloud. In general, the term Digital Twin refers to systems with physical world entities, while in the cyber world exists their digital replicas [9]. This is a widely studied concept and has been applied in fields such as manufacturing [10], informatics [11], robotics [12], aeronautics and space [9]. However, in the automotive industry, only very few works focused on this concept and its applications. In one study [13], a Digital Twin architecture reference model was proposed by Alam et al. for the purpose of developing cloud-based ADAS. A Digital Twin-based framework was developed by Wang et al. for ADAS [14], and a subsequent field implementation of cooperative ramp merging validates its effectiveness with three passenger vehicles [15]. However, one limitation of existing studies is that none of them explores the capability of fusing information

Yongkang Liu and John H.L. Hansen are with CRSS-UTDrive Lab in the Dept. of Electrical and Computer Engineering, University of Texas at Dallas, Richardson, TX 75080, USA (e-mails: {yongkang.liu, john.hansen}@utdallas.edu)

Ziran Wang, Kyungtae Han and Prashant Tiwari are with Toyota Motor North America, InfoTech Labs, Mountain View, CA 94043, USA (e-mails: ryanwang11@hotmail.com, {kyungtae.han, prashant.tiwari}@toyota.com)

Zhenyu Shou is with the Department of Civil Engineering and Engineering Mechanics, Columbia University, New York, NY 10027, USA (e-mail: zs2295@columbia.edu)

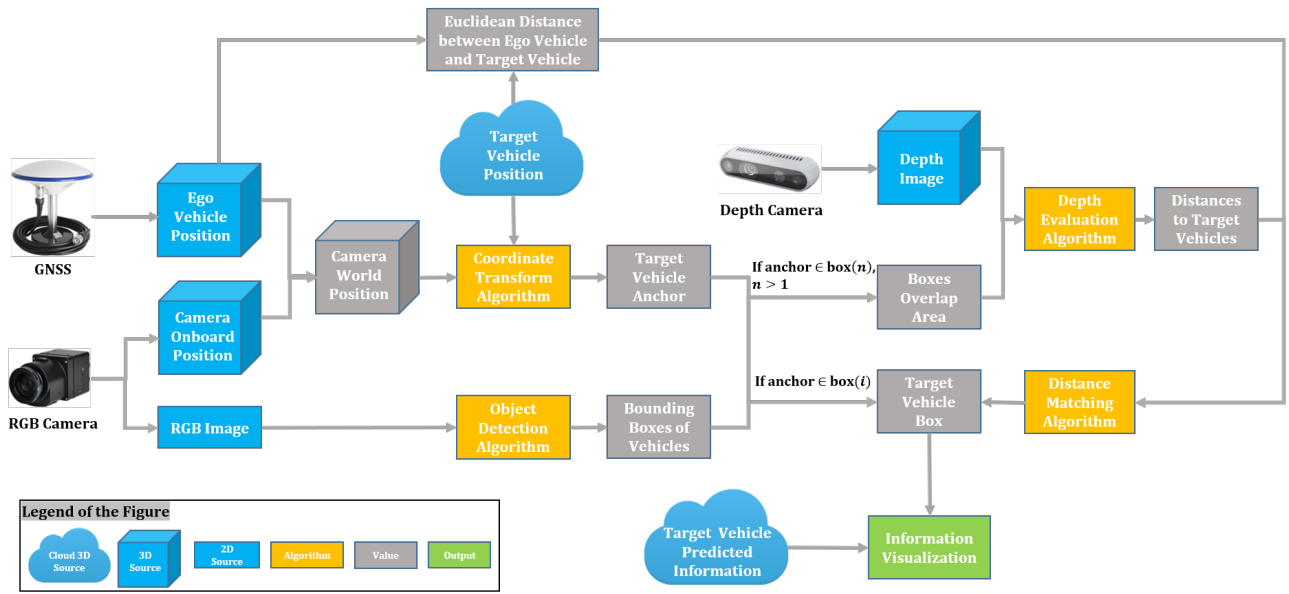

Fig. 1. Proposed vision-cloud data fusion framework. Four key algorithms are included in this architecture. The coordinate transformation algorithm and the object detection algorithm will output the target vehicle anchor point and bounding boxes information regarding target vehicles. The depth evaluation algorithm will provide additional distance knowledge when necessary. This knowledge will assist the distance matching algorithm to identify the correct target vehicle. Finally, predicted information from the cloud will be visualized to provide driver guidance.

provided by the cloud Digital Twin with other sensors' information, which theoretically could yield better prediction of neighboring vehicles behavior, hence provide necessary alerts or favorable guidance to ego-vehicle drivers.

By leveraging Digital Twin technology, cloud servers could create and store digital representations of transportation objects (e.g., pedestrians, vehicles, drivers) in the physical world, then cloud-based modules/algorithms will process such data generated by these entities. Predicted information of target objects (e.g., target vehicle's lane change probability) from cloud servers could be used to augment the perceived information by ego-vehicle's cameras after finishing online processes, hence assist better decision making for the driver or ADAS.

One task of specific interest is the prediction of the lane change maneuver, as the lane change maneuver is an essential driving action that is executed by drivers numerous times every day and has been one of the most common events for accidents. Therefore, we also present lane change prediction as a case study in this work to prove the effectiveness of the proposed data fusion methodology.

The major contributions of this study are listed as follows:
- To the best of the authors' knowledge, this is one of the first studies that explores an effective approach to visualize cloud Digital Twin information to support the decision making of intelligent vehicles.
- The performance comparison between employing single camera source (RGB camera) and employing two camera sources (RGB and depth cameras) for target vehicle identification is studied.
- Human-in-the-loop (HITL) simulation is conducted in a game engine-based intelligent vehicle simulation environment, where safety benefits of implementing this proposed data fusion methodology in a lane change scenario is demonstrated.

The rest of the paper is organized as follows: Sec. II introduces the problem formulation of this study. Sec. III presents the proposed data fusion approach, while the lane change prediction model is presented in Sec. IV. Sec. V discusses the design of the Unity game engine-based simulation and results evaluation. Sec. VI summaries the paper with future study directions.

## II. PROBLEM FORMULATION AND PRELIMINARIES

### A. Problem Statement

One fundamental assumption of this study is that every vehicle participant in the driving scene has access to the internet. Therefore, with vehicle data received by the cloud (e.g., orientation, position, speed, etc.), their Digital Twins (i.e., digital representations of vehicles) can be created on the cloud server. Specifically, localization information can be obtained by the Global Navigation Satellite System (GNSS) onboard, so the statuses of vehicles in the cloud are frequently updated. The lane change prediction model is treated as an online Digital Twin process model, which predicts behavior information regarding the target vehicle (lane change possibility in this study). The predicted information is designed to be visualized back to the driver, assisting the decision making of the intelligent vehicle.

The framework of the proposed approach is presented in Fig. 1. The ego-vehicle has six data sources, including two from the cloud (i.e., target vehicle location and target vehicle predicted information), three on-board 3D sources (i.e., depth image, ego-vehicle location, and onboard camera position), as well as one on-board 2D source (i.e., RGB image). Four key algorithms are included in this architecture: The coordinate transformation algorithm and the object detection algorithm will output the "target vehicle anchor point" and bounding boxes information regarding target vehicles; The depth evaluation algorithm will provide additional distance knowledge when necessary, and this knowledge will assist the distance matching algorithm to identify correct target vehicle. Predicted information from the cloud is then visualized to provide driver guidance.

A key research question is that after receiving the Digital Twin information through the vehicle-to-cloud (V2C) communication, how to correctly overlay such information onto the proper target vehicle. From a secondary viewpoint, one would also need to consider the ego-vehicle's perspective, which motivates a second question on how to identify the target vehicle whose information has been shared [16]. Data obtained by cameras can only be used to detect relatively limited real-time information, while the cloud Digital Twin contains a lot of variables about the target vehicle, especially historical information. To the best of the authors' knowledge, it remains as an open question in the automotive industry on how to properly integrate information from the cloud to complement real-time perception data.

A data fusion approach is proposed in this study to leverage camera information and cloud Digital Twin information, and a lane change prediction case study is conducted. Specifically, as a feature of ADAS on intelligent vehicles, advisory information can be visualized to the driver by utilizing position information obtained from the vehicles' GNSS to identify the target vehicle (with an intention to make the lane change).

*B. Coordinate Transformation*

In this subsection, the system preliminary of coordinate transformation is introduced, which plays an important role in our data fusion methodology. In general, a coordinate transformation algorithm is developed to transform the 3D coordinate of the vehicle (in the world referenced frame) into a 2D point on an image (in the camera referenced frame).

To accomplish this coordinate transformation, a pinhole camera projection model consists of two transformation matrices (extrinsic camera parameter matrix and intrinsic camera parameter matrix) is employed here. Fig. 2 illustrates the model parameters, which are used to describe a 3D point in the physical world and its correspondence 2D projection in the image plane [17].

To build the relation between the camera referenced frame and the world referenced frame, extrinsic parameters are required. Among them, a 3x3 rotation matrix $R$ is used to properly align the corresponding axes of two reference frames. Another component is the translation vector $t$, which describes the relative positions of the origins of these two frames.

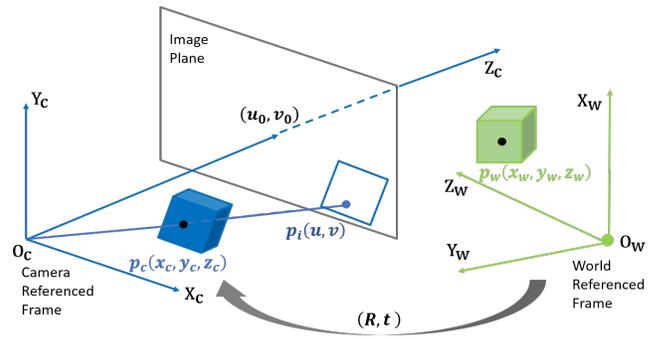

Fig. 2. Transform the 3D world point to its correspondence 2D point in the image plane

Thus, the relationship between a 3D GNSS point $p_w(x_w, y_w, z_w)$ in the world referenced frame $P_w$ and its corresponding point $p_c(x_c, y_c, z_c)$ in the camera referenced frame $P_c$ is

$$p_c = [R|t]\, p_w \qquad (1)$$

Apart from extrinsic parameters, parameters of the camera itself are necessary as well, which are intrinsic parameters. They describe properties inside the camera, such as focal length, lens distortion, and the transformation between image plane and pixels. Depicted in Fig. 2 as well, $(u_0, v_0)$ are the coordinates of the principal point (i.e., the point on the image plane onto which the perspective center is projected) in pixels, $d_x$, $d_y$ are the physical pixel size, and $f$ is the focal length. The intrinsic parameters matrix is therefore

$$M_i = \begin{bmatrix} Z_c d_x/f & 0 & -Z_c d_x u_0/f \\ 0 & Z_c d_y/f & -Z_c d_y v_0/f \\ 0 & 0 & Z_c \end{bmatrix} \qquad (2)$$

The coordinate $p_i(u, v)$ of this GNSS point in the image plane is calculated as

$$p_i = M_i^{-1} p_c \qquad (3)$$

This point will be employed as an "anchor point", which supports future target vehicle identification steps.

*C. Object Detection*

In this subsection, the system preliminary of object detection is introduced, which serves as the key technology in our data fusion methodology. The information provided by the object detection module can be further fused with cloud Digital Twin data to provide visual guidance for the driver.

Generally speaking, the goal of an object detection model is to localize objects with bounding boxes and classify them into different categories or classes. Object detection models have a long-standing history in the computer vision field. Especially in recent years, the development of deep convolutional neural networks [18] largely advanced object detection performance. Nearly all modern object detection models are adopting CNN structures and they generally belong to two groups: two-stage detectors and single-stage detectors. Two-stage detectors will first generate Region of Interests (RoIs) proposals, then use a separate classifier to categorize them in each RoI. Two-stage detectors in general have better performance, but the region proposal stage will yield high computation power cost and slower inference time [19], [20]. Single-stage detectors use a single network to simultaneously detect object positions and predict object classes [21], [22], [23]. Single-stage detectors tend to have fast inference time and low on-board implementation cost, which makes them fit better for real-time applications such as automated driving systems and ADAS.

When developing the driver visual guidance system, the trade-off between accuracy and computational cost needs to be investigated. The object detection module should not only output accurate results but also operate in a timely fashion, so the following modules onboard could have sufficient processing time. Therefore, a single-stage detector is employed in the proposed approach. Since it is not our intention in this study to develop a state-of-the-art object detection algorithm, the object detection module is formulated based on an open source YOLO v3 implementation [24]. The original implementation will output bounding box parameters, confidence score, and classified class information for each detected object. In our methodology, the confidence score is discarded and replaced with a distance value extracted from the corresponding depth image. Then all detection information is saved for the later target vehicle distance matching step, which will be discussed in the next section.

### III. VISION-CLOUD DATA FUSION METHODOLOGY

As illustrated in Fig. 1, to tackle the data fusion problem, four key algorithms are included in this architecture. The coordinate transformation algorithm and the object detection algorithm are introduced as preliminaries in the previous section, which will output the "anchor point" and bounding boxes information regarding target vehicles. The other two algorithms, the depth evaluation algorithm and the distance matching algorithm, will be developed in this section.

*A. Depth Evaluation Algorithm*

Although bounding boxes of target vehicles can be drawn by using the information provided by the object detection model, as Fig. 3 (a) suggested, it is common that detected bounding boxes are not fully separated (i.e., the anchor point is located in multiple bounding box results). Therefore, additional information is desired to help the system accurately identify the target vehicle. Here we propose to use spatial knowledge to assist this process: Fig. 3 (a) is the image taken by an RGB camera and Fig. 3 (b) is the corresponding depth image taken simultaneously by a depth camera. By using this depth image, distance between detected vehicles and the ego-vehicle camera can be computed by our depth evaluation algorithm in Algorithm 1. For each region of the detected bounding box, we randomly sample $n$ points to compute the distance. To avoid mistakenly chosen sample points belonging to the background (e.g., road, sky, etc.) and ensure all samples with good representations, two specific steps are conducted: 1) Truncate the box boundaries by an empirically selected threshold (e.g., with a 0.8 threshold, the new box will be 20% smaller than the original box). The intuition is that a smaller bounding box will most likely be filled by the vehicle itself without road segments; 2) Select points only from the lower quarter area, this is to ensure distance is captured from the backside of detected vehicles to the ego-vehicle, which will provide more accurate distance information.

We set this obtained distance as additional information in our distance matching algorithm, which will be introduced in the next subsection.

*B. Distance Matching Algorithm*

If a single image includes multiple vehicles, the object detection module will output multiple detection results. As the correct information is expected to be overlaid to its corresponding target vehicle properly, a distance matching algorithm is proposed in *Algorithm 2*.

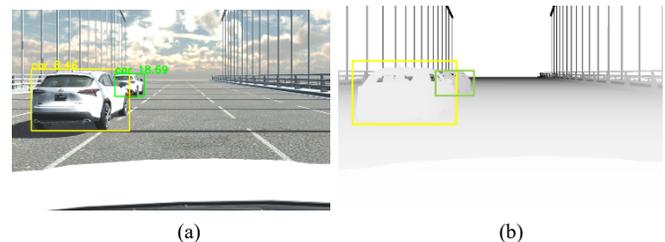

(a) (b)
Fig. 3. (a) One example that detected bounding boxes are overlaid; the green box is the target vehicle. Numbers (18.69 for target vehicle and 8.46 for the rest) are calculated distance results from depth image. This information will assist the target vehicle identification. (b) The corresponding depth image taken simultaneously by a depth camera, used as the input of depth evaluation algorithm.

---

*Algorithm 1*: Depth Evaluation Algorithm

**Input**: Depth image ($Img_d$), detected bounding boxes ($B$), box resize threshold ($th$), total number of sample points ($n$).
**Output**: Distance set ($D$) from detected vehicles to camera.
1:  **for** each detected bounding box $B_i \in B$
2:      decrease the box size according to the box resize threshold ($th$), select the lower ¼ area of the box as $A$;
3:      **for** each point $p_j$, $j < n$
4:          randomly select the point position where $p_j \in A$;
5:          calculate the distance ($pd_j$) between $p_j$ and camera using value from $Img_d$;
6:          put $pd_j$ into the temporary distance set $\Theta$;
7:      **end for**
8:      calculate the distance $d_i = avg(\Theta)$;
9:      put $d_j$ into the distance set $D$;
10: **end for**
11: **return** $D$

---

---

*Algorithm 2*: *Distance Matching Algorithm*

**Input**: Anchor point ($P_i$), detected bounding boxes ($B$), Distance set ($D$) from detected vehicles to camera, distance directly obtained from GNSS ($D_g$).
**Output**: Target vehicle bounding box ($B_t$).
1: **for** each detected bounding box $B_n \in B$
2:     **if** $P_i \in B_n$ **then**
3:        put $B_n$ into the temporary set $\Theta$;
4:        $counter \mathrel{+}= 1$;
5:     **end if**
6: **end for**
7: **if** $counter == 1$ **then**
8:     **return** $B_t = \Theta$
9: **else**
10:     **for** each box $B_j \in \Theta$ and distance $d_j \in D$
11:        calculate the distance difference $\Delta d_j = d_j - D_g$;
12:        put $\Delta d_j$ into the temporary set $T$;
13:     **end for**
14:     find the index $i \in len(T)$ with min(distance difference);
15:     set target box $B_t = \Theta_i$;
16:     **return** $B_t$
17: **end if**

---

Intuitively, based on the anchor point location calculated by the previous coordinate transformation step, we could pick up our target vehicle box if the anchor point is pinpointed exclusively in one detected bounding box. This baseline matching approach works well most of the time but struggles to pick out the correct target vehicle in some corner cases. One example is shown in Fig. 3, where two bounding boxes are overlapped because vehicles are driving in the same lane and close to each other. Because the anchor point location calculated from the previous coordinate transformation step lies in both bounding boxes, the system cannot identify which one is the correct target vehicle. Therefore, spatial knowledge is employed as an additional feature. By comparing the distance obtained from the depth evaluation module with the target vehicle distance communicated from cloud, the candidate box with minimum distance difference is chosen as the target vehicle box.

*C. Methodology Evaluation*

To evaluate the target vehicle identification performance, we conduct an experiment to compare the identification accuracy of our proposed distance matching method (i.e., fusion of RGB camera and depth camera) and the baseline method (i.e., RGB camera only). Various corner cases (e.g., from ego-vehicle driver's viewpoint, two neighboring vehicles are very close to each other) are generated in the simulation environment built by the open source LGSVL simulator [25]. Results are evaluated with the target vehicle identification accuracy across different threshold choices of Intersection over Union (IoU). From Fig. 4, it is easy to tell that the distance matching approach shows better performance than the baseline approach. Spatial knowledge from depth images brings a 7.7% performance improvement (79.2% vs. 73.5%) under the 0.7 IoU threshold. One thing to be noted is that most commonly failed cases come from failed identification of overlapped vehicles, while they are successfully tackled with the help of depth camera. Additionally, the identification accuracy highly depends on the object detection algorithm. In the current implementation, the object detection model is trained on MSCOCO dataset [26], and a higher accuracy is expected after fine tuning the object detection model with dataset designed for autonomous driving.

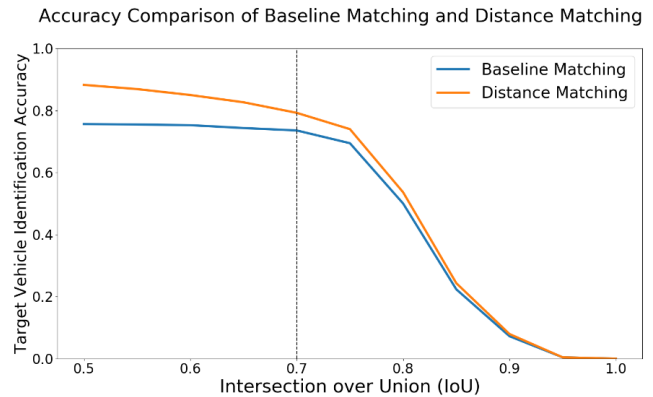

Fig. 4. Target vehicle identification accuracy-IoU curve

## IV. CASE STUDY ON LANE CHANGE PREDICTION

*A. Lane Change Maneuver and Driving Safety*

As shown in the system architecture Fig. 1, the cloud Digital Twin is supposed to provide the target vehicle predicted information, so this result can be visualized to the driver via our vision-cloud data fusion methodology. In this section, we specifically conduct a case study on the lane change prediction task, which is an option to implement our proposed methodology as ADAS on mass-produced vehicles.

Lane change maneuver is an essential driving action executed millions of times every day and is one of the riskiest maneuvers that a driver must perform in a conventional highway system. Therefore, we propose a long-term (i.e., 5~10 seconds) prediction model of the lane change maneuver to improve driving safety.

Numerous efforts have been made on the lane change prediction task over the past few years [27], [28], [29]. Most previous studies employ lateral or steering wheel angle information as features, enabling only a relatively short prediction horizon (i.e., < 5 seconds). If a vehicle's lateral position is approaching the lane mark, or the steering angle moves to point towards an adjacent lane, the possibility of that vehicle making a lane change is high. However, in this case study, we make long-term lane change prediction without using lateral or steering angle information of the vehicle.

*B. Multi-Layer Perceptron for Lane Change Prediction*

A three-layer multilayer perceptron (MLP) classifier is utilized in our lane change prediction model [30]. Features we explored include speed of the ego-vehicle, speed difference between the ego-vehicle and its neighboring vehicles, and longitudinal gap among them. Although it is relatively easy to extract lane change maneuvers from data, there is no clear clue on how much time it takes for a driver to prepare before the lane change maneuver is executed. Therefore, a time-window labeling method is adopted to make features from positive and negative samples more distinguishable.

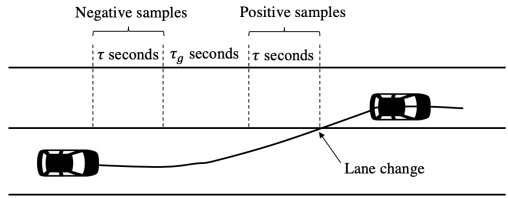

Fig. 5. A time-window labeling method

*1) Time-Window Labeling:* The time-window labeling method employed in this case study is shown in Fig. 5. Data within a time-window of $\tau$ seconds at the lane change end point is labeled as positive samples. Negative samples are obtained by moving the time-window $\tau + \tau_g$ seconds backward. The difference in this method is that, a gap of $\tau_g$ between positive and negative samples is set, other than directly labeling samples at $\tau$ seconds ahead of positive samples as negative (i.e., $\tau_g = 0$). The intuition is that the driving environment, in general, will not change dramatically during a short time. If $\tau_g = 0$, the last few frames in negative samples and the first few frames in positive samples will contain very similar features but assigned to different labels. This will introduce difficulties in the model training phase Therefore, by adding a gap of $\tau_g$ between them, we expect that features from negative samples and positive samples will be more distinguishable. To balance the number of positive and negative samples, we adopted the same time-window size for them, which is critical for training our prediction model.

*2) Modified Prediction Methodology:* One issue encountered during run-time evaluation is the occurrence of unstable predictions. A lane change prediction example is shown in TABLE I, where 0 and 1 mean second-by-second predictions of the target vehicle. A left lane change was made at timestep 8, where our MLP model predicts the vehicle to perform lane change at timesteps 4 and 8. However, this prediction is unstable in the sense that the positive predictions are not consistent. Although the MLP prediction in Table I does alert the driver to be cautious at timestep 4, it suggests no precaution during timestep 5-7, which only brings confusing from the vehicle driver's perspective. To improve this unstable prediction situation, two alternative approaches, namely an aggressive approach and a conservative approach, are proposed to further refine the lane change prediction model.

TABLE I
A SAMPLE PERIOD OF LANE CHANGE GROUND TRUTH AND PREDICTION, WHERE 0 MEANS STAY PUT AND 1 MEANS LANE CHANGE

| Timestep | Ground Truth | Original MLP Prediction | Aggressive MLP Prediction | Conservative MLP Prediction |
|---|---|---|---|---|
| 1 | 0 | 0 | 0 | 0 |
| 2 | 0 | 0 | 0 | 0 |
| 3 | 0 | 0 | 0 | 0 |
| 4 | 0 | 1 | 1 | 0 |
| 5 | 0 | 0 | 1 | 0 |
| 6 | 0 | 0 | 1 | 0 |
| 7 | 0 | 0 | 1 | 0 |
| 8 | 1 | 1 | 1 | 0 |

The aggressive approach is presented as *Algorithm 3*. It propagates a positive prediction for additional $\tau_a$ timesteps. Instead of only last 1 timestep, now each positive prediction lasts for $(\tau_a + 1)$ timesteps. Take the fourth column of TABLE I as an example, it represents the results after applying the aggressive approach to the prediction in the third column with $\tau_a = 3$.

Alternatively, Algorithm 4 presents the conservative approach. It smooths out the prediction results through an average fashion. At each timestep, we take the average of the current prediction and predictions for the previous $\tau_c$ timesteps (i.e., the average of these $(\tau_c + 1)$ values). If the result exceeds the threshold, then we make the prediction at this frame as positive. Otherwise, we assign this as negative. For example, predictions are 1 at timestep 4 and 0 at previous $\tau_c = 3$ timesteps. The average is thus 0.25 which is below the threshold 0.5 used in this study. The prediction at timestep 4 is therefore modified from 1 to 0.

After using either the aggressive or conservative approach, the prediction results are less fluctuating. However, limitations exist in both approaches: More positive predictions will be introduced in the aggressive approach, resulting in a higher false positive rate (i.e., more false alarms); While for the conservative approach, it suppresses some positive predictions, leading to a lower true positive rate (i.e., lower prediction

---

*Algorithm 3: Aggressive-approach MLP Prediction*
**Input**: Lane change predictions $lc_t$ from the model at each frame $t \in \{1,2,\cdots,T\}$, parameter $\tau_a$.
**Output**: Modified lane change predictions $\widehat{lc}_t$
01: Initialize $\widehat{lc}_t$ as zeros
02: **for** $t \in \{1,2,\cdots,T\}$ **do**
03:   **if** $lc_t = 1$ **then**
04:     **for** $\tau \in \{0,1,\cdots,\tau_a\}$ **do**
05:       **if** $t + \tau \leq T$ **then**
06:         $\widehat{lc}_{t+\tau} = 1$
07:       **end if**
08:     **end for**
09:   **end if**
10: **end for**
11: **return** $\widehat{lc}_t$

---

*Algorithm 4: Conservative-approach MLP Prediction*
**Input**: Lane change predictions $lc_t$ from the model at each frame $t \in \{1,2,\cdots,T\}$, parameter $\tau_c$, and threshold value *thres*
**Output**: Modified lane change predictions $\widehat{lc}_t$
01: Initialize $\widehat{lc}_t$ as zeros
02: **for** $t \in \{\tau_c, \tau_c + 1, \cdots, T\}$ **do**
03:   $s = 0$
04:   **for** $\tau \in \{0,1,\cdots,\tau_c\}$ **do**
05:     $s \mathrel{+}= lc_{t-\tau}$
06:   **end for**
07:   $avg = s/(\tau_c + 1)$
08:   **if** $avg > thres$ **then**
09:     $\widehat{lc}_t = 1$
10:   **end if**
11: **end for**
12: **return** $\widehat{lc}_t$

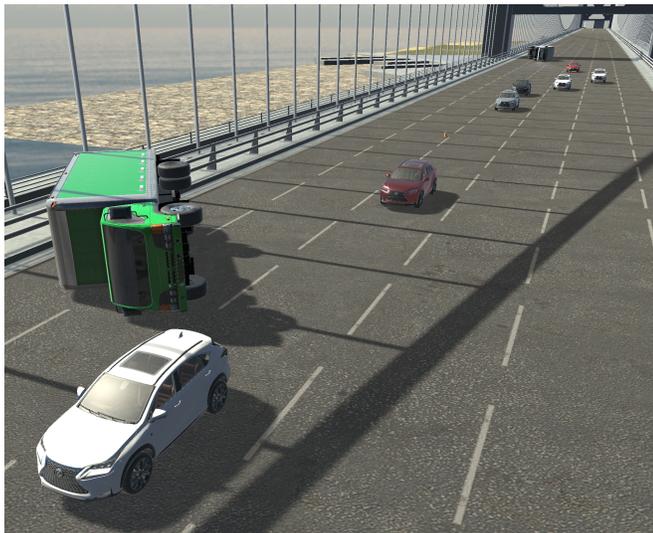

Fig. 6. Multi-lane highway environment built in Unity, where the furthest red vehicle is our ego-vehicle

accuracy). Thus, the tradeoff needs to be considered, and both parameters ($\tau_a$ and $\tau_c$) need to be carefully adjusted depending on application scenarios. In this case study, the aggressive MLP model is adopted to minimize the false negative rate and ensure driving safety.

## V. GAME ENGINE SIMULATION AND RESULTS EVALUATION

### A. Game Engine Simulation in Unity

Naturalistic driving data is essential to provide better understanding of driver behaviors [8], [31]. However, it is time-consuming and expensive for data collection platform implementation, naturalistic driving data collection, and human annotation. On the other hand, the advantage of easy implementation of large-scale traffic environment, and its flexibility to collect enormous data under various and sometimes dangerous driving scenarios makes game engine-based simulation much helpful at the early stage of system development.

Game engines (such as Unity [32] and Unreal [33]) are software systems that consist of a graphics rendering engine, a physical engine (for object movement, interaction, and collision detection), and an UI interference for managing elements (e.g., sound tracks, models, visual effects, etc.). Game engine-based test environments were explored by various studies to prototype connected vehicles [34], [35], model driver behaviors [36], and simulate autonomous driving [25], [37]. In this study, Unity is adopted to conduct modeling and evaluation of the proposed data fusion system, given its strengths in visualization, graphics design, as well as external joystick (i.e., driving simulator) integration.

### B. Human-in-the-Loop (HITL) Simulation Setup

In this simulation, a multi-lane highway environment is built in Unity, illustrated in 0. An accident scene is designed in this environment, where two stopped trucks occupy the two right lanes (along their forward direction). When vehicles are approaching this accident zone, some of them might make left lane change actions to avoid potential collision, and in turn create conflicts to the motion trajectory of the ego-vehicle (the furthest red vehicle). Our proposed lane change prediction model is therefore supposed to provide prediction information to the ego-vehicle's driver.

TABLE II
PARAMETER SETUP OF THE HITL SIMULATION IN UNITY GAME ENGINE

| Parameters | Value |
|---|---|
| Time step of simulation update | 0.01 s |
| Time step of vehicle-to-cloud communication in simulation | 0.1 s |
| Times step of AR information visualization update | 0.1 s |
| Distance traveled by vehicles in the simulation | 300 m |
| Number of neighboring vehicles | 6 |
| Number of potential lane-change neighboring vehicles | 3 |
| Initial speed of the ego-vehicle | 19 m/s |
| Initial speed of all neighboring vehicles | 17 m/s |

Regarding vision-cloud data fusion module, the ego-vehicle is implemented with both RGB and depth cameras in Unity, where RGB images could be generated as Fig. 3(a) and depth images could be generated as Fig. 3(b). Real-time vehicle positions are obtained by a GNSS module as a 3D coordinate in the same environment. Since we assume all vehicles are intelligent vehicles with V2C communication ability in this environment, the ego-vehicle can retrieve neighboring vehicles' predicted information through Unity C# API. An Augmented Reality (AR)-based information visualization approach is implemented in Unity to display the predicted lane change probability to the driver as a head-up display of the ego-vehicle.

The HITL simulation platform is built with a laptop (Intel Core i7-9750 @2.60 GHz processor, 32.0 GB memory, NVIDIA Quadro RTX 5000 graphic card), an external 1920×1080 full HD resolution 23.8-inch monitor, a Logitech G29 Driving Force racing wheel, and the Unity game engine 2019.2.11f1. In this simulation, two invited participants (with experience on the simulation platform but different levels of real-world driving skills) are requested to drive the ego-vehicle through this accident scene in two separate scenarios, each takes around 30 seconds. In one scenario, AR guidance, generated by the proposed data fusion methodology and lane change prediction model, is provided to the ego-vehicle's driver. The other scenario works as the baseline, where no guidance information is provided at all, so the ego-vehicle is driven as a legacy vehicle. The parameter setup of this HITL simulation is included in Table II.

### C. Evaluation Results

Comprehensive HITL simulation studies have been conducted by participants for both scenarios (with and without using the proposed system). Two example simulation periods are picked out in to illustrate the effectiveness of the proposed system. As can be seen in Fig. 7(a1)-(a4) where the proposed system is implemented, the AR guidance is overlaid on top of the silver target vehicle based on the data fusion methodology. The AR guidance information is color-coded, where the visualization color is dynamically associated to the predicted lane change probability. The color is yellow when the probability is medium, and red when the probability is high. In this example simulation trip, since the driver was aware of the potential lane change at an early stage, the vehicle was able to decelerate in advance to keep a safe gap with the silver lane-change vehicle.

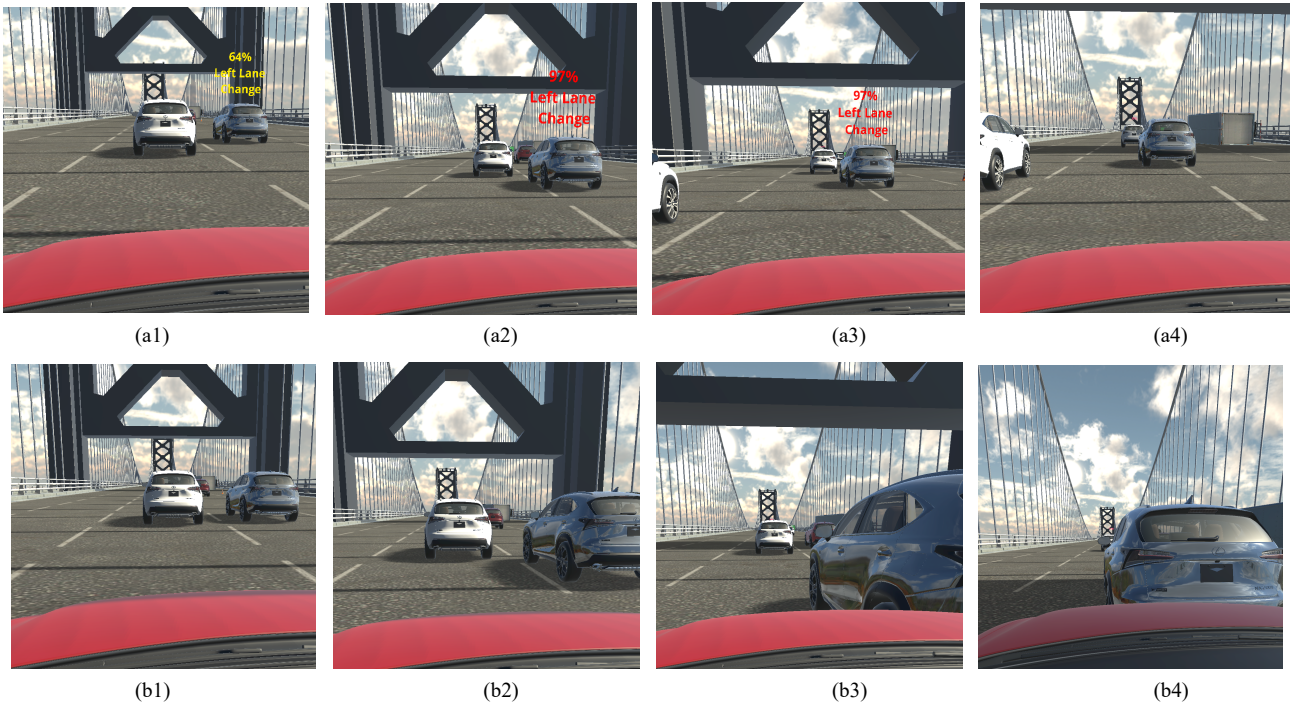

Fig. 7. Comparison in human-in-the-loop simulation between scenario a) when the proposed system is implemented (a1–a4), and baseline scenario b) without the proposed system (b1–b4). With the proposed system, the driver was able to decelerate in advance to keep a safe gap with the silver lane-change vehicle. Without the proposed system, the driver decelerated at a very late stage due to the sudden lane-change behavior of the silver vehicle, and the ego-vehicle nearly collided with the lane-change vehicle.

In the baseline scenario, no guidance information is visualized to the driver at all as shown in Fig. 7(b1)-(b4). Since the silver neighboring vehicle suddenly changed the lane to the left, the ego-vehicle's driver had very limited time to react, hence it nearly collided with that neighboring vehicle. Note that the initial positions of all neighboring vehicles are randomly generated in each trip, so participants of the HITL simulation did not drive the exact same scenario repeatedly, and hence in the baseline scenario, like Fig. 7(b1)-(b4), they were not aware of any lane-change decision of neighboring vehicles in advance.

We compare both scenarios (with and without use of the proposed system) in terms of driving safety and comfort. Three different measurement factors are compared in all HITL simulation trips in both scenarios: 1) average time-to-collision (TTC) between the ego-vehicle and the silver lane-change vehicle, 2) average absolute acceleration of the ego-vehicle, and 3) maximum jerk of the ego-vehicle during the trip. The results are shown in Fig. 8, where a notable increase of average TTC value can be seen after implementing the proposed model. The increase of 30.31% means the driver of the ego-vehicle tended to keep further away from the lane-change vehicle when the visual guidance was provided. It is suggested that a larger TTC value can induce safer driving performance, and prevent potential rear-end collisions in such emergency situations.

It is also visible from Fig. 8 that the average absolute acceleration decreases from 2.57 m/s$^2$ to 2.07 m/s$^2$ after the proposed vision-cloud data fusion model is implemented/deployed. In general, if a driver conducts acceleration and deceleration maneuvers more frequently, the accident rate will also increase due to the large speed variance. With lane change prediction information visualized to the driver, he/she can drive in a milder manner with precautions regarding future situations.

Additionally, a decrease of 10.54% in the maximum jerk value was reached using the proposed model. Jerk is defined as the first derivative of vehicle acceleration and it reflects the comfort level of driving maneuvers. The decrease of jerk from 40.14 m/s$^3$ to 35.91 m/s$^3$ indicates that the proposed model can introduce a more comfortable and smooth driving/riding experience for the driver/passengers.

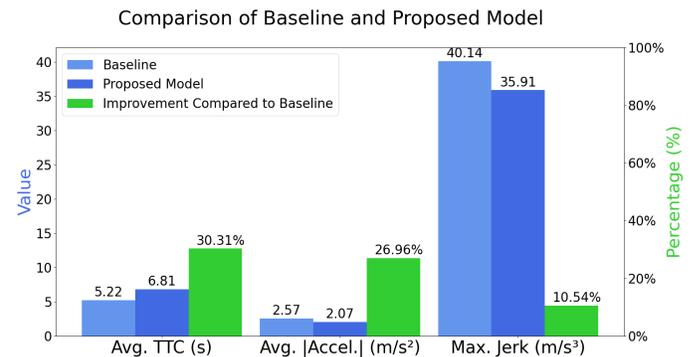

Fig. 8. Comparison results in the HITL simulation in terms of safety and comfort measurements

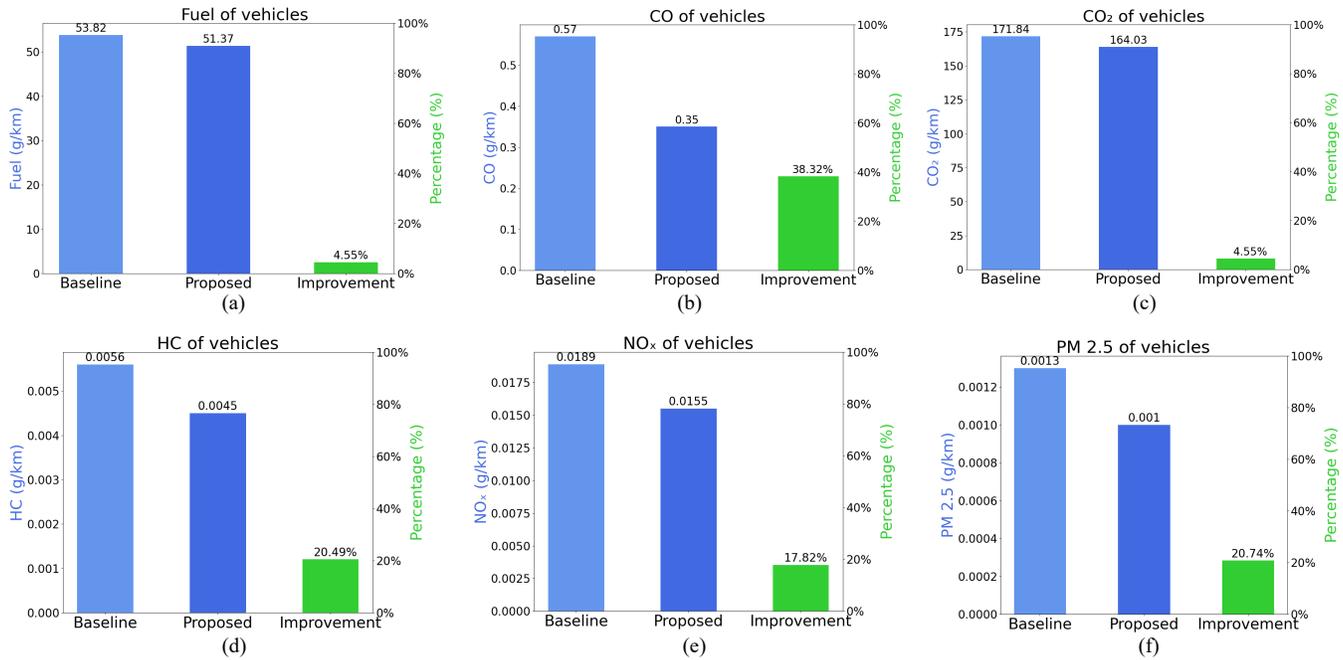

Fig. 9. Comparison results in the HITL simulation in terms of fuel and emission measurements

In addition to safety and comfort perspectives, this study also evaluates the environmental benefits that our proposed model could bring to current transportation systems. The open-source MOVESTAR model is adopted to calculate fuel consumption and pollutant emissions generated by the ego-vehicle [38]. As shown in Fig. 9 (a), after implementing the proposed model, the average fuel consumption changed from 53.82 g/km to 51.37 g/km, achieving an overall 4.55% reduction. Based on Fig. 9 (b)-(f), various sources of pollutant emissions were also reduced compared to baseline, when the proposed vision-cloud data fusion model (with lane change prediction information) was provided to the driver. Specifically, carbon monoxide had the greatest level of reduction with a 38.32% change.

All these reductions in fuel consumption as well as pollutant emissions can be considered as byproducts of reducing the average absolute acceleration, since less speed changes intuitively lead to less fuel consumption and emissions produced by the vehicle.

## VI. CONCLUSIONS AND FUTURE WORK

In this study, a vision-cloud data fusion framework was proposed, with an overall aim to visualize cloud Digital Twin information to the driver in an intelligent vehicle. A target vehicle identification strategy utilizing spatial information was also explored, achieving an overall 79.2% accuracy under 0.7 IoU threshold. A case study on a realistic ADAS application was also conducted, where an MLP algorithm with modified prediction approaches proposed to conduct lane change prediction. HITL simulation in the Unity game engine revealed that the proposed model brings benefits to highway driving in terms of safety, comfort, and environmental sustainability.

Since there may have latencies in communication and update frequency limitation in GNSS information, future studies could consider integrating location estimation algorithm into the system to maintain the target vehicle identification accuracy, as well as fine-tune the model with the real-world data to improve performance. Another potential future research direction is to test the proposed visualization system in real-world passenger vehicles, potentially with a separate center console display or an advanced head-up display device. When considering future field implementation, it is also important to explore physical constraints such as the minimum refresh interval requirement of GNSS data, and time clock matching for different sensors.

ACKNOWLEDGMENT

The contents of this paper reflect only the views of the authors, who are responsible for the facts and accuracy of the data presented herein. The content does not necessarily reflect the official views of Toyota Motor North America.

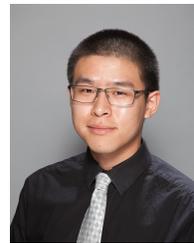

**Yongkang Liu** received the Ph.D. and M.S. degrees in electrical engineering from the University of Texas at Dallas in 2021 and 2017, and the B.S. degree in electronic information engineering from Shandong Normal University in 2015, respectively. He is currently a Research Engineer at Toyota Motor North America, InfoTech Labs. His research interests are focused on in-vehicle systems, advancements in smart vehicle technologies, and driver behavior modeling.

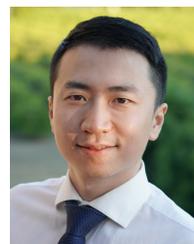

**Ziran Wang** (S'16-M'19) received the Ph.D. degree in mechanical engineering from the University of California at Riverside in 2019, and the B.E. degree in mechanical engineering and automation from Beijing University of Posts and Telecommunications in 2015, respectively. He is currently a Principal Researcher at Toyota Motor North America, InfoTech Labs. He is an associate editor of SAE International Journal of Connected and Automated Vehicles, founding chair of the IEEE ITSS technical committee on Internet of Things in Intelligent Transportation Systems, and member of four other technical committees across IEEE and SAE. Dr. Wang's research focuses on intelligent vehicles, including cooperative automation, human factors, and Digital Twin.

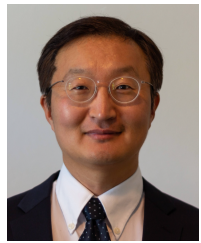

**Kyungtae (KT) Han** (M'97-SM'15) received the Ph.D. degree in electrical and computer engineering from The University of Texas at Austin in 2006. He is currently a Principal Researcher at Toyota Motor North America, InfoTech Labs. Prior to joining Toyota, Dr. Han was a Research Scientist at Intel Labs, and a Director in Locix Inc. His research interests include cyber-physical systems, connected and automated vehicle technique, and intelligent transportation systems.

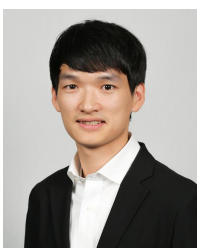

**Zhenyu Shou** received the Ph.D. degree in civil engineering from Columbia University in 2020, and the B.S. degree in theoretical and applied mechanics from Peking University in 2015, respectively. He was a research intern at Toyota Motor North America R&D, InfoTech Labs, and a quantitative strategy research intern at Pinebridge Investment. His research interests include driver behavior modeling and resource allocation.

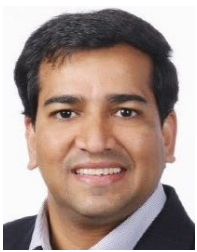

**Prashant Tiwari** (M'20) received the Ph.D. degree in mechanical engineering from Rensselaer Polytechnic Institute in 2004, and the MBA degree from University of Chicago in 2016. He is currently an Executive Director at Toyota Motor North America, InfoTech Labs. Dr. Tiwari's Division is responsible for performing applied research for future connected technologies with focus on car-to-car, car-to-edge and car-to-cloud connectivity on bean end-to-end basis and building Toyota's next generation mobility service platform and edge computing capability. Dr. Tiwari is highly active in Automotive Edge Computing Consortium (AECC) and SAE. Prior to joining Toyota, Dr. Tiwari held several leadership positions of increasing responsibilities at GE & UTC. In his last role with GE, Dr. Tiwari served as Senior Director at GE Aviation Digital. Prior to joining Toyota, Dr. Tiwari held Executive position and led the Embedded Systems & Integrated Project Teams for UTC Aerospace Systems.

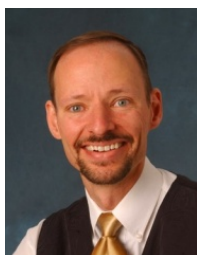

**John H.L. Hansen** (S'81-M'82-SM'93-F'07) received Ph.D. and M.S. degrees in electrical engineering from Georgia Institute of Technology, and B.S.E.E. degree from Rutgers University. He joined Univ. of Texas-Dallas (UTDallas) in 2005, where he is currently Associate Dean for Research, and previously served as Dept. Head of Electrical Engineering. He is an IEEE Fellow and past elected SLTC Chair and Member of IEEE Signal Processing Society Speech-Language Technical Committee, and Educational Technical Committee. His research interests span the areas of digital speech processing, machine learning for speech and speaker traits, feature estimation in noise, cognitive modeling and driver distraction analysis for effective human-machine interactions. At UT-Dallas, Hansen established the Center for Robust Speech Systems (CRSS), and oversees the UTDrive Lab, which is focused on driver modeling and distraction.